\documentclass{article}

\usepackage[preprint]{corl_2025} 

\usepackage{xcolor}
\usepackage{amssymb}
\usepackage{amsmath}
\usepackage{graphicx}
\usepackage{booktabs}
\usepackage{pifont}
\usepackage{caption}
\usepackage{wrapfig}
\usepackage{enumitem}

\newcommand{\method}{\textsc{ViTaL}}
\newcommand{\methodbc}{\textsc{ViTaL-BC}}
\newcommand{\website}{\href{https://vitalprecise.github.io/}{vitalprecise.github.io}} 

\title{Touch begins where vision ends: \\ Generalizable policies for contact-rich manipulation}

\newcommand{\xxnote}[3]{}
\ifx\hidenotes\undefined
  \renewcommand{\xxnote}[3]{\color{#2}{#1: #3}}
\fi

\definecolor{olivegreen}{HTML}{3C8031}
\author{
Zifan Zhao$^{1}$ \qquad Siddhant Haldar$^{2}$ \qquad Jinda Cui$^{3}$ \qquad Lerrel Pinto$^{2}$ \qquad Raunaq Bhirangi$^{2}$\thanks{Correspondence to: raunaqbhirangi@nyu.edu} \\
\\
$^{1}$New York University Shanghai \qquad
$^{2}$New York University \qquad $^{3}$Honda Research \\
\\
{\small \tt \website{}}
\vspace{-0.1in}
}

\makeatletter
\let\@oldmaketitle\@maketitle%
\renewcommand{\@maketitle}{\@oldmaketitle%
    \centering
    \includegraphics[width=0.9\linewidth]{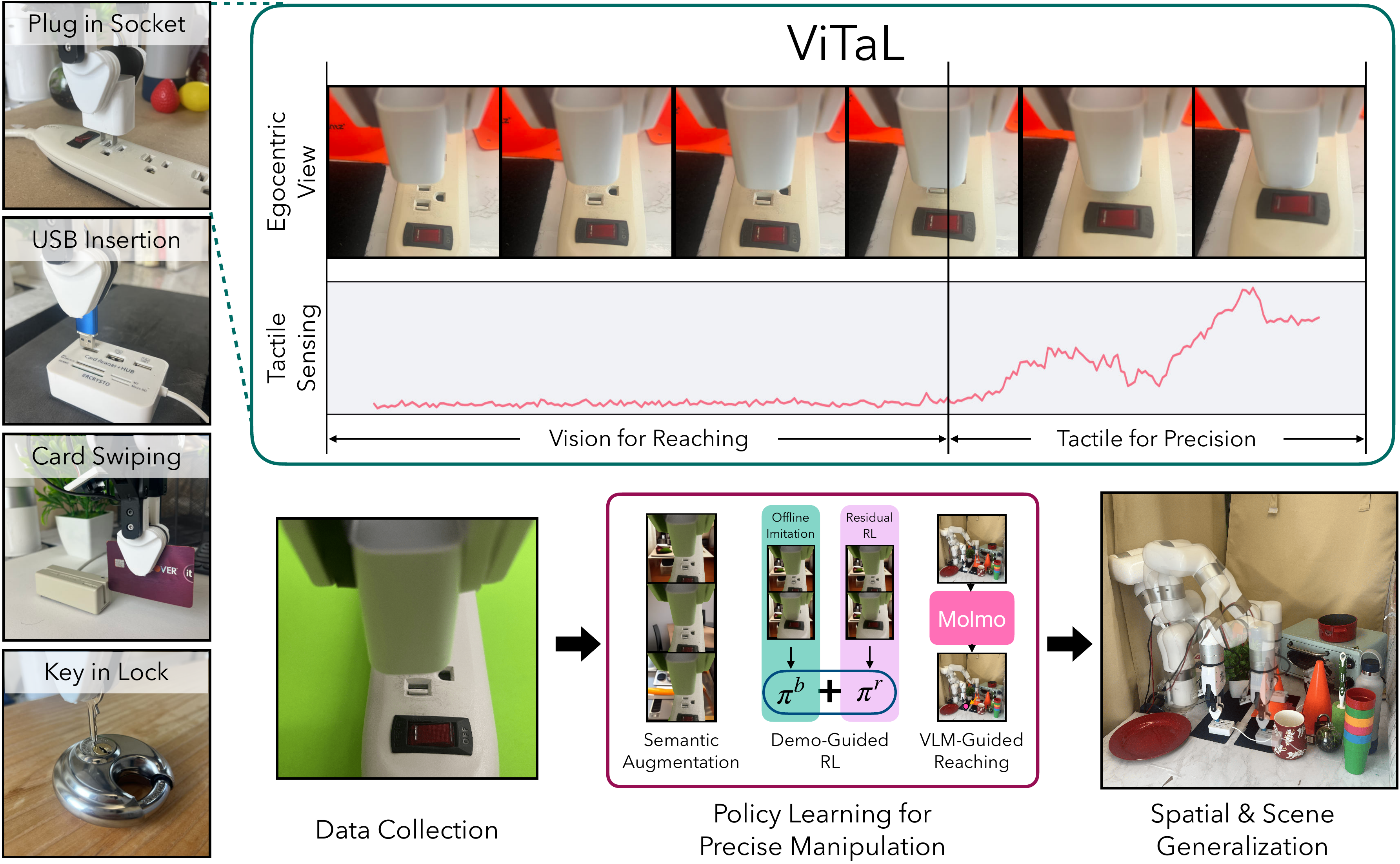}
    \captionof{figure}{We present \method{}, a framework that combines tactile sensing, vision foundation models, and residual reinforcement learning to enable learning policies that can operate with millimeter-level precision while generalizing to large spatial variations and significant environmental perturbations.}
    \label{fig:intro}
}
\makeatother

\begin{document}
\maketitle


\begin{abstract}

    Data-driven approaches struggle with precise manipulation: imitation learning requires many hard-to-obtain demonstrations, while reinforcement learning yields brittle, non-generalizable policies. We introduce VisuoTactile Local (\method{}) policy learning, a framework that solves fine-grained manipulation tasks by decomposing them into two phases: a \textit{reaching phase}, where a vision-language model (VLM) enables scene-level reasoning to localize the object of interest, and a local \textit{interaction phase}, where a reusable, scene-agnostic \method{} policy performs contact-rich manipulation using egocentric vision and tactile sensing. This approach is motivated by the observation that while scene context varies, the low-level interaction remains consistent across task instances. By training local policies once in a canonical setting, they can generalize via a localize-then-execute strategy. \method{} achieves $\sim$90\% success on contact-rich tasks in unseen environments and is robust to distractors. \method{}’s effectiveness stems from three key insights: (1) foundation models for segmentation enable training robust visual encoders via behavior cloning; (2) these encoders improve the generalizability of policies learned using residual RL; and (3) tactile sensing significantly boosts performance in contact-rich tasks. Ablation studies validate each of these insights, and we demonstrate that \method{} integrates well with high-level VLMs, enabling robust, reusable low-level skills. Results and videos are available at \website.

\end{abstract}


\section{Introduction}
\label{sec:introduction}

Imitation learning for sensorimotor skills has made significant strides in recent years, propelled by the increasing scale and diversity of robotic datasets. From controlled tabletop environments to open-world household settings, large-scale data has been shown to improve the generalization of robotic policies~\cite{etukuru2024robot, black2024pi0} -- mirroring advances in vision~\cite{dalle3,imagen,sam2} and language~\cite{gpt4,gemini,llama}. However, precise, contact-rich manipulation poses a significant challenge to this data-centric approach. Fine-grained tasks such as inserting USBs and swiping credit cards have low error tolerance (millimeter to sub-millimeter), and the high fidelity required makes demonstration collection time-consuming, brittle, and difficult to scale. Deep reinforcement learning (RL) provides an alternative by learning directly through online interaction, but often sacrifices generalization in favor of narrowly tuned policies sensitive to training-specific cues like scene layout or background distractors. 

In this work, we propose VisuoTactile Local (\method{}), a policy learning framework that bridges this gap by enabling robust, precise manipulation while maintaining generalizability. \method{} decomposes manipulation into two phases: a global \textit{reaching phase}, where a vision-language model (VLM) performs scene-level reasoning to identify and localize the object of interest, and a local \textit{interaction phase}, where a reusable, scene-agnostic policy performs fine-grained, contact-rich manipulation using egocentric vision and tactile sensing. This decomposition is motivated by the observation that while the environmental context for a task may vary drastically, the low-level physical interactions required for manipulation remain consistent. Our work focuses on capturing this invariant local policy: training it once in a canonical setting allows it to generalize across environments via a simple localize-then-execute strategy. With just 32 demonstrations and 45 minutes of online reinforcement learning per task, \method{} achieves the precision necessary for real-world deployment while maintaining adaptability across scenes.

A core design motivation behind \method{} is the deliberate pairing of sensing modalities with complementary strengths. Tactile sensing is indispensable during contact-rich phases of manipulation, providing direct, localized feedback about forces and slip, that cannot be captured by vision. It is inherently robust to lighting, background clutter, and occlusion, but lacks the spatial awareness necessary for planning and coarse alignment in the pre-contact phase. Egocentric vision fills this gap by offering a consistent, robot-centered perspective that captures the relative pose of the end-effector and surrounding objects. Unlike third-person or fixed external cameras, egocentric views are naturally aligned with the robot’s actions and are easy to replicate across different environments without introducing viewpoint-specific biases that can severely hinder learned policy transfer.

While visuotactile design is not novel in itself, existing works typically fail to use it effectively. Imitation learning methods require large, diverse datasets~\cite{levine2016endtoendtrainingdeepvisuomotor, rt22023arxiv} to handle spatial and scene variation, making them expensive and difficult to scale, especially for precise manipulation. Reinforcement learning is capable of refining policies through interaction, but tends to overfit to training environments~\cite{cobbe2019quantifyinggeneralizationreinforcementlearning,zhang2018studyoverfittingdeepreinforcement}. A key reason for this is that learning from raw RGB inputs in constrained settings lacks the visual diversity needed for generalization. Without sufficient variation in appearance, background, and lighting, policies trained via RL become brittle and environment-specific.

\method{} addresses this limitation with a key insight: task success depends primarily on the visual features of task-relevant objects, which remain relatively stable across environmental changes. To exploit this invariance, we introduce a semantic, task-aware data augmentation pipeline powered by vision foundation models. These augmentations introduce altering distractors, backgrounds, and lighting, while preserving object and robot identity. This allows visual encoders to learn more general representations from the same amount of demonstration data, eliminating the need for costly scene variations in data collection.

Finally, to further improve performance and address the inevitable imperfections in teleoperated demonstrations, we fine-tune our policies using offset-based reinforcement learning. Rather than learning policies from scratch, we apply DrQ-v2~\cite{drqv2} to refine behavior-cloned policies by predicting small corrective actions, or offsets, relative to the predicted actions. Crucially, this refinement is done without discarding the visual generalization learned during imitation, as we continue to apply semantic augmentations during online training. This final phase boosts precision and robustness while preserving the broad generalization enabled by our visuotactile design and augmentation strategy.

Our key findings can be summarized as follows:
\begin{enumerate}[leftmargin=*,align=left]
    \item \method{} learns generalizable, contact-rich manipulation policies with a 90\% success rate from just 32 demonstrations and 45 minutes of interaction, outperforming the best baseline by 40\% on average across four challenging precise manipulation tasks in unseen environments.
    \item Tactile sensing is essential for precision and reliability: removing tactile input reduces success rates by an average of 40\%, underscoring its critical role in contact-rich task phases where vision alone is insufficient.
    \item \method{} extends the benefits of semantic visual augmentation beyond imitation learning by combining it with residual RL, enabling policy fine-tuning without sacrificing generalization.
    
\end{enumerate}

All of our datasets, and training and evaluation code have been made publicly available. Videos of our trained policies can be seen here:~\website{}.

\section{Related Work}

\subsection{Imitation and Reinforcement Learning}
Unlike language modeling and computer vision, robot learning has been heavily constrained by limited data \cite{lin2024data}. Imitation learning, often from teleoperated expert demonstrations, has gained momentum in recent years as a method for transferring human manipulation skills to robots \cite{lee2024behavior, Zhao2023act, chi2023diffusionpolicy}. Although the quality of demonstration data has steadily improved, its quantity remains limited, making it difficult for learned policies to handle corner cases or generalize to unseen variations \cite{lin2024data, chi2024universal}. This challenge persists even with Vision-Language-Action models (VLAs), which combine large pretrained models with imitation learning-based fine-tuning \cite{black2024pi0,li2024cogact, wen2024tinyvla}.  In contrast, reinforcement learning (RL) is known for its ability to generalize, driven by exploration and domain randomization \cite{kroemer2021review, tobin2017domain}. However, RL typically suffers from poor data efficiency and sim2real gaps \cite{andrychowicz2020learning,zhao2020sim}. Online residual RL has emerged as a promising solution, combining the strengths of both approaches: it uses a base policy to guide exploration for better efficiency while simultaneously refining the policy \cite{haldar2023fish, guzey2024tavi, yuan2024policy, silver2018residual, johannink2019residual,zhang2019deep,alakuijala2021residual}. Building on this insight, our method adopts residual RL as its foundation. 

\subsection{Precise Manipulation}
Beyond pick-and-place operations, robots must be capable of physically interacting with and modifying their environments using objects as tools. Many such tasks require precise control over contact interactions \cite{billard2019trends, suomalainen2022survey, hogan2020tactile}. Tactile sensing is critical for this purpose, as vision alone often fails to capture subtle deformations at contact points and can be hindered by occlusions. Traditionally, researchers have relied on model-based analytical methods for contact-rich tasks \cite{whitney1986mechanical,wu20241khz,pang2023global}, but these approaches tend to lack robustness or extendability in unstructured environments. Learning-based methods, on the other hand, are often tailored to specific tasks \cite{beltran2020variable, zhao2022offline, qi2023hand}, or struggle to effectively integrate tactile sensing with vision for high-precision manipulation \cite{guzey2023dexterity,guzey2024tavi,lin2024learning}. In this work, we present a scalable approach to contact-rich tasks by leveraging local tactile sensing to constrain and guide contact interactions, enabling reliable and precise task execution.

\subsection{Priors for Generalizable Policy Learning}
Generalization efficiency is particularly crucial in robot learning due to the scarcity of real-world data. To address this challenge, researchers have introduced various forms of inductive bias into learning frameworks. These priors are often infused at the representation level: common computer vision techniques are typically used to augment visual inputs in robot learning \cite{xu2023comprehensive}, while alternative approaches leverage object-centric properties, such as equivariance, to further enhance the performance \cite{simeonov2022neural, wang2022robot}. Moreover, the generalizability of pretrained models is often harnessed to enable downstream tasks \cite{florence2018dense, morgan2022complex,levy2024p3,haldar2025point}. More generally, invariance in manipulation tasks has been exploited to retarget or generate trajectories that cover a wider range of variations \cite{mandlekar2023mimicgendatagenerationscalable, johns2021coarse}. In our work, beyond extensive background visual augmentation, we build a local sensing structure and employ task-space control to enable natural generalization to spatial variations.
\section{\method{}}

\begin{figure}[t]
    \centering
    \includegraphics[width=\linewidth]{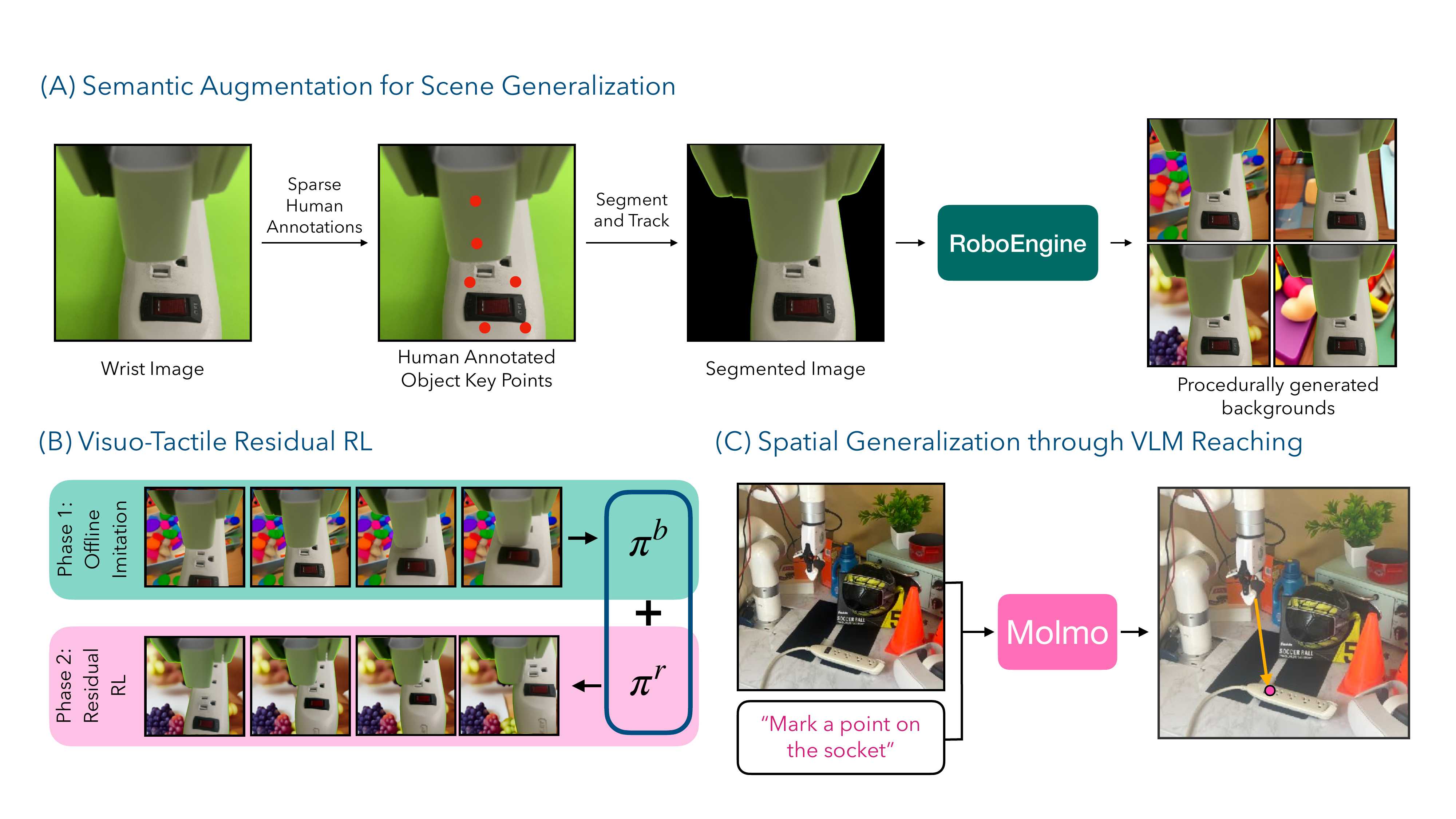}
    \caption{Overview of \method{}. (A) \method{} utilizes vision foundation models to enhance task data with procedurally generated backgrounds, improving visual diversity. (B) This data is then used to train a generalizable visuo-tactile policy, which is later refined through online residual reinforcement learning (RL) for precision. (C) Finally, VLM-guided reaching enables zero-shot deployment in novel spatial configurations, despite policies being trained on fixed object positions.}
    \label{fig:method}
    \vspace{-1em}
\end{figure}

The core insight behind \method{} is that vision offers task-level spatial awareness for scene generalization, while tactile sensing is essential for millimeter-scale precision during physical contact. By leveraging the strength of each modality, our method enables policies to be trained in localized settings and deployed across diverse spatial variations and background configurations. \method{} operates in three phases: $(1)$ Visuotactile behavior cloning learns a generalizable base policy using visual semantic augmentations; $(2)$ Residual RL enhances downstream performance by optimizing policy refinements while maintaining vision-driven robustness; $(3)$ VLM-based reaching facilitates zero-shot adaptation to novel spatial configurations by identifying actionable regions and decoupling task dynamics from environment configuration. Our pipeline has been illustrated in Figure~\ref{fig:method}.


\subsection{Generalizable behavior cloning through semantic augmentations}
\label{sec:bcsecaug}
Our method starts by collecting visuotactile robot demonstrations using a virtual reality (VR) based teleoperation framework~\cite{openteach}. All the tasks presented in this paper consist of a target object on the table that the robot interacts with, and a grasped object that is held within the robot gripper. We hypothesize that for most precise manipulation tasks, the core interaction dynamics remain consistent across task instances, despite variations in the broader environment, ie., the dynamics of plugging your charger in the kitchen are consistent with the dynamics of plugging your charger in the bedroom. To focus data collection on these invariant interactions, we fix the target object position, and collect successful demonstrations with the robot randomly initialized in the vicinity of the target object. We ensure that observations and actions are grounded in the robot's end-effector frame to enable transfer to novel spatial configurations during inference. This is achieved by using the wrist camera image and tactile readings as input observations, and computing relative actions in the robot's end effector frame. By constraining spatial variability and focusing on local interaction patterns, our method achieves robust policy learning with only $32$ demonstrations per task.

To maintain policy performance across variations in the visual environment, we implement semantic augmentations targeting visual regions irrelevant to the task. Our collected demonstrations use a green screen background to facilitate background replacement through procedural scene generation~\cite{greenaug} using RoboEngine~\cite{roboengine} during policy learning. In our initial experiments, we observed that naive color-key based background filtering performs poorly, which prompted our multi-stage segmentation pipeline: First, a human annotator marks key points on task-relevant objects in a \textit{single} reference demonstration frame. This often requires only a few seconds. Next, DIFT-based correspondence matching~\cite{dift} propagates these annotations to the first frame of all demonstrations, followed by Segment-Anything 2~\cite{sam2} for instance segmentation. Finally, XMem~\cite{xmem} tracks the segmented masks temporally along trajectories, separating the relevant task elements from augmentable background regions (Fig.~\ref{fig:method}). This allows targeted background transformations while preserving contact-relevant visual features critical for tactile coordination. 

The demonstration data is then used to train a base visuotactile policy using behavior cloning. The augmented visual data is encoded using a randomly-initialized ResNet-18~\cite{resnet} encoder, and tactile reading from AnySkin~\cite{anyskin} is encoded using a multilayer perception (MLP). The encoded observations are fed into a visuo-tactile transformer policy $\pi^b$ for action prediction~\cite{visk}. The policy is trained with action chunking~\cite{Zhao2023act} using a mean squared error loss between predicted and ground truth action chunks. By jointly enforcing spatial and visual invariance through semantic augmentations and sensory observations grounded in the end-effector frame, the policy develops robust task understanding decoupled from environmental context.

%

\subsection{Fine-tuning with Demonstration-guided Reinforcement Learning}

While the pretrained base policy $\pi^b$ enables generalizable visuo-tactile policies, we observe that the policy only achieves a modest success rate. To improve the performance of $\pi^b$, we employ residual reinforcement learning (RL) to train a residual policy $\pi^r$ on top of the base policy. In residual RL~\cite{silver2018residual}, given a base policy $\pi^b: \mathcal{Z}\rightarrow\mathcal{A}$ with encoded representations $z\in\mathcal{Z}$ and action $a\in\mathcal{A}$, we learn a residual policy $\pi^r: \mathcal{Z}\times\mathcal{A}\rightarrow\mathcal{A}$ such that an action sampled from the final policy $\pi$ is the sum of the base action $a^b\sim\pi^{b}(z)$ and the residual offset $a^r\sim\pi^{r}(z, a^b)$. Following prior work~\cite{,haldar2023fish,guzey2024tavi}, we use n-step DDPG~\cite{ddpg} as our RL optimizer, a deterministic actor-critic based method that provides high performance in continuous control~\cite{drqv2}. 

During online learning, the encoders (ResNet-18 for vision, MLP for tactile) trained for behavior cloning are fixed, and feed compressed representations $z^i$ (image) and $z^t$ (tactile) to both the frozen base policy $\pi_b$ and the residual actor network, $\pi_r$, which takes as input $(z^i, z^t, a^b)$ to predict $a^r$. Similarly, the residual critic $Q^r$ evaluates $(z^i, z^t, a^b, a^r)$ pairs using layer normalization and high update-to-data (UTD) ratios for sample-efficient Q-learning~\cite{smith2022walk}. 
Crucially, we observe that adding  L2 weight regularization for the actor network improves policy training, resulting in better performance. For RL training, our reward is simply a sum of a binary success reward provided by a human at the end of the trajectory and a dense L1 distance from the goal for the task. The RL training objective is as follows:

\begin{equation}
\pi^{r} = \operatorname*{argmax}_{\pi^{r}} \mathbb{E}_{(z^i, z^t, a^b,a^r)\sim\mathcal{D}_{\beta}}\left[Q(z^i,z^t,a^b,a^r)\right] 
\label{eq:residual}
\end{equation}

where $\mathcal{D}_{\beta}$ contains rollouts enriched with the same semantic visual augmentations from the behavior cloning phase to maintain generalization.
The executed action $a$ is a sum of $a^b$ and $a^r$. This approach of combining fixed pretrained features with adaptive residuals improves policy performance while preserving cross-environment robustness through augmentations. Details about hyperparameters and network architectures used in our experiments have been included in Appendix~\ref{app:hyperparams}.







\subsection{Inference}

Our framework achieves spatial and scene generalization through a hierarchical inference strategy: global semantic navigation by a high-level agent followed by localized visuotactile control for precise low-level execution. By combining offline behavior cloning and online residual adaptation, the policy operates within a constrained task space while maintaining robustness to environmental perturbations. For global positioning, we employ Molmo~\cite{molmo}, a vision-language model (VLM) pretrained on web-scale data, to coarsely localize target objects specified via natural language. 

Given an external RGB-D observation, Molmo predicts a 2D coordinate for the target object, which is projected to 3D workspace coordinates using depth data and camera calibration parameters. The robot then samples an initial end-effector pose within a pre-defined region of the target coordinate. For example, for USB insertion, the target point for the robot is sampled at a height of 10cm above the predicted coordinate. Empirically, we observe that this coarse initialization falls within the pretrained policy’s operational envelope, ensuring target visibility in the wrist camera feed. Upon reaching the target position, the learned visuotactile local policy is deployed to complete the task. Our results in Section~\ref{sec:experiments} demonstrate the potential of combining general-purpose VLMs for coarse robotic navigation, with localized visuo-tactile policies handling the precise parts of a task.

\section{Experiments}
\label{sec:experiments}

Our experiments seek to answer the following questions: $(1)$ How does \method{} perform in an in-domain setting? $(2)$ How does \method{} perform under environmental perturbations?  $(3)$ What are the important design choices for \method{}? $(4)$ How well does the VLM navigation work with \method{}?

\subsection{Experimental Setup}

Our experiments are conducted using a UFACTORY xArm 7 robot equipped with a two-fingered xArm Gripper. For tactile sensing, we integrate the AnySkin~\cite{anyskin} magnetic tactile sensor into the gripper. The observations for policy learning include $128 \times 128$ RGB images captured by a fisheye camera mounted on the robot’s wrist, and 15-dimensional tactile readings from the AnySkin sensor. For coarse navigation via the VLM, we use a calibrated third-person Intel RealSense RGB-D camera. For each task, demonstrations are collected using a VR-based teleoperation system~\cite{openteach} operating at 30 Hz. The collected data is subsampled to 6Hz for policy training, and the learned policies are deployed at 6Hz during real-world execution.

\subsection{Task Descriptions}
\label{subsec:task_desc}

We demonstrate the versatility of our framework by evaluating \method{} on four precise, contact-rich manipulation tasks and a pick bread task. We collect 32 demonstrations for each task while fixing the target object and randomly initializing the robot in a predefined area around it. Detailed task descriptions can be found in Appendix~\ref{app:task-descriptions}.

\subsection{Baselines}
We compare \method{} with four primary baselines -- BAKU~\cite{baku}, ViSk~\cite{visk}, and RLPD~\cite{rlpd}, and the semantically augmented BC policy described in Section~\ref{sec:bcsecaug}. 

\noindent\textbf{BAKU~\cite{baku}:} Transformer policy for behavior cloning that maps RGB images to robot actions.

\noindent\textbf{ViSk~\cite{visk}:} BAKU with both RGB images and tactile readings provided as input.

\noindent\textbf{RLPD~\cite{rlpd}:} This involves collecting a few expert demonstrations and training an RL policy from scratch, where the data during RL training is sampled 1:1 between the expert and RL replay buffer.

\noindent\textbf{\methodbc{}:} Visuotactile base policy described above that uses the semantic augmentation scheme with the ViSk architecture.

Further details on baseline implementations can be found in Appendix~\ref{app:baseline}.



\begin{table*}[t]
\centering
\caption{Policy performance of \method{} in an in-domain setting.}
\label{table:indomain}
\renewcommand{\tabcolsep}{4pt}
\renewcommand{\arraystretch}{1}
\begin{tabular}{lccccc}
\toprule
\multicolumn{1}{c}{\textbf{Method}} & \textbf{\begin{tabular}[c]{@{}c@{}}Plug in\\ Socket\end{tabular}} & \textbf{\begin{tabular}[c]{@{}c@{}}Insert\\ USB\end{tabular}} & \textbf{\begin{tabular}[c]{@{}c@{}}Card\\ Swiping\end{tabular}} & \textbf{\begin{tabular}[c]{@{}c@{}}Key in\\ Lock\end{tabular}} & \textbf{\begin{tabular}[c]{@{}c@{}}Pick\\ Bread\end{tabular}} \\
\midrule
BAKU~\cite{baku} & 4/10 & 4/10 & 1/10 &  5/10 & \textbf{10/10}\\
ViSK~\cite{visk} & 7/10 & 3/10 & 4/10 & 5/10 & \textbf{10/10}\\
RLPD~\cite{rlpd} & 3/10 & 0/10 & 2/10 & 1/10 & \textbf{10/10}\\
\methodbc{} & 7/10 & 4/10 & 5/10 & 5/10 & \textbf{10/10} \\
\method{} \textbf{(Ours)} & \textbf{9/10} & \textbf{9/10} & \textbf{10/10} & \textbf{9/10} & \textbf{10/10} \\
\bottomrule
\end{tabular}
\end{table*}

\subsection{How does \method{} perform in an in-domain setting?}
Table~\ref{table:indomain} evaluates \method{}'s performance in a controlled in-domain setting, where both the background (green screen) and object positions are fixed. For each method, we conduct 10 trials per task, with the robot randomly initialized within a predefined area around the target object (Section~\ref{subsec:task_desc}). Both \method{} and RLPD receive identical visual and tactile observations and are trained online for 45 minutes. While \method{} incorporates semantic augmentations in its RL replay buffer, we find that such augmentations degrade performance for RLPD; therefore, RLPD results in Table~\ref{table:indomain} do not use semantic augmentations. Our results demonstrate that \method{} significantly outperforms all baselines, achieving an absolute improvement of 40\% over the strongest alternative. Notably, ViSk outperforms BAKU, highlighting the importance of tactile sensing for precise manipulation. Further, \method{} surpasses RLPD, emphasizing the value of offline policy pretraining for sample-efficient online learning. Overall, these findings illustrate that visuotactile behavior cloning and residual RL scaffolded by semantic augmentations enables robust, high-precision manipulation.

\begin{table*}[t]
\centering
\caption{Study of spatial and scene generalization in \method{}.}
\label{table:generalization}
\renewcommand{\tabcolsep}{4pt}
\renewcommand{\arraystretch}{1}
\begin{tabular}{lcccccccc}
\toprule
\multicolumn{1}{c}{\textbf{Method}} & \multicolumn{4}{c}{\textbf{Spatial Generalization}}                                                                                                                                                            & \multicolumn{4}{c}{\textbf{Scene Generalization}}                                                                                                                                                         \\ 
\cmidrule(lr){2-5}
\cmidrule(lr){6-9} & \textbf{\begin{tabular}[c]{@{}c@{}}Plug in \\ Socket\end{tabular}} & \textbf{\begin{tabular}[c]{@{}c@{}}Insert \\ USB\end{tabular}} & \textbf{\begin{tabular}[c]{@{}c@{}}Card \\ Swiping\end{tabular}} & \textbf{\begin{tabular}[c]{@{}c@{}}Key in \\ Lock\end{tabular}} & \textbf{\begin{tabular}[c]{@{}c@{}}Plug in \\ Socket\end{tabular}} & \textbf{\begin{tabular}[c]{@{}c@{}}Insert \\ USB\end{tabular}} & \textbf{\begin{tabular}[c]{@{}c@{}}Card \\ Swiping\end{tabular}} & \textbf{\begin{tabular}[c]{@{}c@{}}Key in \\ Lock\end{tabular}} \\ \midrule
BAKU~\cite{baku}   & 15/30 & 5/30 & 7/30 & 11/30 & 0/30 & 0/30 & 0/30 & 0/30 \\
ViSK~\cite{visk}   & 19/30 & 6/30 & 15/30 & 14/30 & 0/30 & 0/30 & 0/30 & 0/30 \\
RLPD~\cite{rlpd}   &  5/30 & 0/30 & 3/30 & 2/30 & 0/30 & 0/30 & 0/30 & 0/30 \\
\methodbc{}        & 21/30 & 10/30 & 16/30&  12/30 & 24/30 & 8/30 & 13/30&  17/30 \\
\method{}          & \textbf{28/30} & \textbf{24/30} & \textbf{28/30} & \textbf{22/30} & \textbf{27/30} & \textbf{23/30} & \textbf{25/30} & \textbf{24/30}\\ \bottomrule
\end{tabular}
\end{table*}

\begin{table*}[t]
\caption{Study of \method{}'s robustness to combined spatial and scene perturbations.}
\label{table:spatial_scene_gen}
\centering
\begin{tabular}{lcccc}
\toprule
\multicolumn{1}{c}{\textbf{Method}} & \textbf{\begin{tabular}[c]{@{}c@{}}Plug in\\ Socket\end{tabular}} & \textbf{\begin{tabular}[c]{@{}c@{}}Insert\\ USB\end{tabular}} & \textbf{\begin{tabular}[c]{@{}c@{}}Card\\ Swiping\end{tabular}} & \textbf{\begin{tabular}[c]{@{}c@{}}Key in\\ Lock\end{tabular}} \\
\midrule
BAKU~\cite{baku} & 0/30 & 0/30 & 0/30 &  0/30\\
ViSK~\cite{visk} & 0/30 & 0/30 & 0/30 & 0/30\\
RLPD~\cite{rlpd} & 0/30 & 0/30 & 0/30 & 0/30 \\
\methodbc{} & 24/30 & 9/30 & 19/30 & 13/10 \\
\method{} \textbf{(Ours)} & \textbf{29/30} & \textbf{25/30} & \textbf{27/30} & \textbf{27/30}\\
\bottomrule
\end{tabular}
\end{table*}

\subsection{How does \method{} perform under environmental perturbations?}
\label{subsec:scenegen}

\paragraph{Spatial Generalization} Table~\ref{table:generalization} evaluates \method{}'s spatial generalization by testing three novel target object positions outside the training distribution, with the green screen background retained to isolate spatial variations from scene-level changes. Across 10 trials per position, each initializing the robot within a predefined workspace around the target object, results show comparable performance to in-domain settings, confirming that localized end-effector frame observations effectively enable spatial generalization. Notably, BAKU and ViSk admit a performance decline when target objects approach the edges of the green screen, resulting in background elements entering into the fisheye wrist camera’s field of view, inducing visual distribution shifts relative to training data.

\paragraph{Scene Generalization} Table~\ref{table:generalization} assesses \method{}’s scene generalization by testing on three novel, cluttered scene configurations (see Appendix~\ref{app:scenegen} for examples) while keeping the target object position fixed and identical to training. For each configuration, we run 10 trials with the robot randomly initialized within a predefined area around the target. The results demonstrate \method{}’s robustness to unstructured scene variations, significantly outperforming all baselines. The strong performance of both \method{} and \methodbc{} highlights the critical role of semantic augmentations in enabling policies to disentangle task-relevant visual cues from environmental noise. Moreover, \method{}’s improvement over \methodbc{} illustrates how residual RL combined with semantic augmentations substantially enhances performance while preserving \methodbc{}’s generality. Table~\ref{table:spatial_scene_gen} extends this evaluation to scenarios varying both target spatial positions and background appearances. To decouple policy performance from VLM navigation effects, we manually initialize the robot near the target object and conduct 10 trials per position. The results revealing a consistent pattern: \method{} and \methodbc{} outperform baselines, with \method{} maintaining a clear advantage. Overall, the use of localized observation spaces alongside semantic augmentations during training endows \method{} with strong spatial and scene generalization capabilities.

\subsection{What are the important design choices for \method{}?}
\label{subsec:ablations}
\method{} is an amalgam of several techniques that enable learning generalizable visuo-tactile policies. Here, we systematically ablate several design choices in \method{} and justify their importance. 

\begin{table*}[htbp]
\caption{Study of important design choices for \method{}.}
\label{table:ablation_tactile}
\centering
\begin{tabular}{lccc}
\toprule
\multicolumn{1}{c}{\textbf{Method}} & \textbf{\begin{tabular}[c]{@{}c@{}}Plug in\\ Socket\end{tabular}} & \textbf{\begin{tabular}[c]{@{}c@{}}USB\\ Insertion\end{tabular}} & \textbf{\begin{tabular}[c]{@{}c@{}}Card\\ Swiping\end{tabular}} \\ 
\midrule
\textit{Tactile Ablations} \\
Visual BC & 4/10 & 4/10 & 1/10 \\
Visuo-Tactile BC & 7/10 & 3/10 & 4/10 \\
Visual BC + Res. RL & \textbf{9/10} & 7/10 & 0/10 \\
\midrule
\textit{Semantic Augmentation Ablations} \\
Visual BC (BAKU) & 0/10 & 0/10 & 0/10 \\
Visual BC + Aug. & 4.7/10 & 2.3/10 & 0.7/10 \\
Visuo-Tactile BC (ViSk) & 0/10 & 0/10 & 0/10 \\
Visuo-Tactile BC + Aug. & 8.3/10 & 3/10 & 6.3/10 \\
\midrule
\method{} & \textbf{9/10} & \textbf{9/10} & \textbf{10/10} \\
\bottomrule
\end{tabular}
\end{table*}

\paragraph{Tactile sensing} Table~\ref{table:ablation_tactile} investigates tactile sensing's role in enabling millimeter-scale precision, with experiments conducted under controlled conditions (fixed object positions, green screen background) to isolate sensory effects. Comparing visual (BAKU) and visuo-tactile (ViSk) BC, both with and without residual RL, reveals a consistent performance advantage with tactile inputs. While visual BC with residual RL is competent on two tasks, utilizing tactile inputs further improves performance. Qualitatively, this improvement stems from visual occlusion challenges: as the end effector approaches the target, the object held by the gripper obstructs the egocentric camera’s view of the goal, rendering visual feedback unreliable and causing hesitation or blind actions. Tactile sensing proves indispensable in tasks like \textit{Card Swiping}, where the card occludes the machine and the policy has to heavily rely on tactile sensing for task completion. The results confirm that tactile sensing compensates for dynamic visual obstructions while enabling finer contact-driven adjustments.

\paragraph{Semantic Augmentation} Table~\ref{table:ablation_tactile} studies the importance of semantic augmentations for novel scene generalization. We average the performance of visual (BAKU) and visuo-tactile (ViSK) BC -- with and without semantic augmentations --  across three unseen object positions with background distractors. Our results demonstrate that semantic augmentations enable both approaches to adapt to new spatial and visual conditions, with visuotactile BC achieving superior performance than its visual counterpart.

\subsection{How well does the VLM navigation work with \method{}?}
\begin{wraptable}{l}{7.4cm}
\vspace{-1em}
\caption{VLM Navigation for spatial generalization.}
\label{table:vlm}
\begin{tabular}{lcccc}
\toprule
\multicolumn{1}{c}{\textbf{Method}} & \textbf{\begin{tabular}[c]{@{}c@{}}Plug in\\ Socket\end{tabular}} & \textbf{\begin{tabular}[c]{@{}c@{}}USB\\ Insertion\end{tabular}} & \textbf{\begin{tabular}[c]{@{}c@{}}Card\\ Swiping\end{tabular}} \\ 
\midrule
ViSk~\cite{visk} & 0/25 & 0/25 & 0/25 \\
\methodbc{} & 16/25 & 9/25 & 13/25 \\
\method{} \textbf{(Ours)} & 21/25 & 16/25 & 19/25 \\
\bottomrule
\end{tabular}
\vspace{-1em}
\end{wraptable}

Table~\ref{table:vlm} evaluates VLM-based coarse navigation across five novel object positions, conducting five trials per position while including background distractors to test robustness to environmental perturbations. Compared to the strongest baseline, \methodbc{}, we observe that both methods generalize to unseen object positions and maintain consistent performance in cluttered scenes, despite being trained on fixed configurations. This highlights the utility of VLM navigation for imparting spatial robustness to visuotactile policies.

\section{Conclusion and Limitations}
\label{sec:conclusion}


This work introduces \method{}, a framework that integrates local observation spaces and semantic augmentations for visuotactile policy learning with VLM-guided coarse navigation to achieve millimeter-precision manipulation across diverse scenes and object configurations. We recognize a few limitations of this work: (1) Our spatial variation experiments focus on horizontal surfaces -- extending the method to arbitrary 3D configurations (e.g., vertical or angled placements) would be an interesting future direction. (2) Our VLM navigation assumes obstacle-free paths. Enhancing spatial reasoning to handle cluttered environments through advanced VLMs with obstacle-aware path planning could broaden the applicability of the method. (3) Our current evaluations use controlled lab conditions. Testing in real-world home environments with dynamic lighting, occlusions, and unstructured layouts would better validate robustness.

\bibliography{references}  

\begin{thebibliography}{65}
\providecommand{\natexlab}[1]{#1}
\providecommand{\url}[1]{\texttt{#1}}
\expandafter\ifx\csname urlstyle\endcsname\relax
  \providecommand{\doi}[1]{doi: #1}\else
  \providecommand{\doi}{doi: \begingroup \urlstyle{rm}\Url}\fi

\bibitem[Etukuru et~al.(2024)Etukuru, Naka, Hu, Lee, Mehu, Edsinger, Paxton, Chintala, Pinto, and Shafiullah]{etukuru2024robot}
H.~Etukuru, N.~Naka, Z.~Hu, S.~Lee, J.~Mehu, A.~Edsinger, C.~Paxton, S.~Chintala, L.~Pinto, and N.~M.~M. Shafiullah.
\newblock Robot utility models: General policies for zero-shot deployment in new environments.
\newblock \emph{arXiv preprint arXiv:2409.05865}, 2024.

\bibitem[Black et~al.(2024)Black, Brown, Driess, Esmail, Equi, Finn, Fusai, Groom, Hausman, Ichter, Jakubczak, Jones, Ke, Levine, Li-Bell, Mothukuri, Nair, Pertsch, Shi, Tanner, Vuong, Walling, Wang, and Zhilinsky]{black2024pi0}
K.~Black, N.~Brown, D.~Driess, A.~Esmail, M.~Equi, C.~Finn, N.~Fusai, L.~Groom, K.~Hausman, B.~Ichter, S.~Jakubczak, T.~Jones, L.~Ke, S.~Levine, A.~Li-Bell, M.~Mothukuri, S.~Nair, K.~Pertsch, L.~X. Shi, J.~Tanner, Q.~Vuong, A.~Walling, H.~Wang, and U.~Zhilinsky.
\newblock $\pi_0$: A vision-language-action flow model for general robot control, 2024.
\newblock URL \url{https://arxiv.org/abs/2410.24164}.

\bibitem[Betker et~al.(2023)Betker, Goh, Jing, Brooks, Wang, Li, Ouyang, Zhuang, Lee, Guo, et~al.]{dalle3}
J.~Betker, G.~Goh, L.~Jing, T.~Brooks, J.~Wang, L.~Li, L.~Ouyang, J.~Zhuang, J.~Lee, Y.~Guo, et~al.
\newblock Improving image generation with better captions.
\newblock \emph{Computer Science. https://cdn. openai. com/papers/dall-e-3. pdf}, 2\penalty0 (3):\penalty0 8, 2023.

\bibitem[Saharia et~al.(2022)Saharia, Chan, Saxena, Li, Whang, Denton, Ghasemipour, Gontijo~Lopes, Karagol~Ayan, Salimans, et~al.]{imagen}
C.~Saharia, W.~Chan, S.~Saxena, L.~Li, J.~Whang, E.~L. Denton, K.~Ghasemipour, R.~Gontijo~Lopes, B.~Karagol~Ayan, T.~Salimans, et~al.
\newblock Photorealistic text-to-image diffusion models with deep language understanding.
\newblock \emph{Advances in neural information processing systems}, 35:\penalty0 36479--36494, 2022.

\bibitem[Ravi et~al.(2024)Ravi, Gabeur, Hu, Hu, Ryali, Ma, Khedr, R{\"a}dle, Rolland, Gustafson, Mintun, Pan, Alwala, Carion, Wu, Girshick, Doll{\'a}r, and Feichtenhofer]{sam2}
N.~Ravi, V.~Gabeur, Y.-T. Hu, R.~Hu, C.~Ryali, T.~Ma, H.~Khedr, R.~R{\"a}dle, C.~Rolland, L.~Gustafson, E.~Mintun, J.~Pan, K.~V. Alwala, N.~Carion, C.-Y. Wu, R.~Girshick, P.~Doll{\'a}r, and C.~Feichtenhofer.
\newblock Sam 2: Segment anything in images and videos.
\newblock \emph{arXiv preprint arXiv:2408.00714}, 2024.
\newblock URL \url{https://arxiv.org/abs/2408.00714}.

\bibitem[Achiam et~al.(2023)Achiam, Adler, Agarwal, Ahmad, Akkaya, Aleman, Almeida, Altenschmidt, Altman, Anadkat, et~al.]{gpt4}
J.~Achiam, S.~Adler, S.~Agarwal, L.~Ahmad, I.~Akkaya, F.~L. Aleman, D.~Almeida, J.~Altenschmidt, S.~Altman, S.~Anadkat, et~al.
\newblock Gpt-4 technical report.
\newblock \emph{arXiv preprint arXiv:2303.08774}, 2023.

\bibitem[Reid et~al.(2024)Reid, Savinov, Teplyashin, Lepikhin, Lillicrap, Alayrac, Soricut, Lazaridou, Firat, Schrittwieser, et~al.]{gemini}
M.~Reid, N.~Savinov, D.~Teplyashin, D.~Lepikhin, T.~Lillicrap, J.-b. Alayrac, R.~Soricut, A.~Lazaridou, O.~Firat, J.~Schrittwieser, et~al.
\newblock Gemini 1.5: Unlocking multimodal understanding across millions of tokens of context.
\newblock \emph{arXiv preprint arXiv:2403.05530}, 2024.

\bibitem[Touvron et~al.(2023)Touvron, Lavril, Izacard, Martinet, Lachaux, Lacroix, Rozi{\`e}re, Goyal, Hambro, Azhar, et~al.]{llama}
H.~Touvron, T.~Lavril, G.~Izacard, X.~Martinet, M.-A. Lachaux, T.~Lacroix, B.~Rozi{\`e}re, N.~Goyal, E.~Hambro, F.~Azhar, et~al.
\newblock Llama: Open and efficient foundation language models.
\newblock \emph{arXiv preprint arXiv:2302.13971}, 2023.

\bibitem[Levine et~al.(2016)Levine, Finn, Darrell, and Abbeel]{levine2016endtoendtrainingdeepvisuomotor}
S.~Levine, C.~Finn, T.~Darrell, and P.~Abbeel.
\newblock End-to-end training of deep visuomotor policies, 2016.
\newblock URL \url{https://arxiv.org/abs/1504.00702}.

\bibitem[Brohan et~al.(2023)Brohan, Brown, Carbajal, Chebotar, Chen, Choromanski, Ding, Driess, Dubey, Finn, Florence, Fu, Arenas, Gopalakrishnan, Han, Hausman, Herzog, Hsu, Ichter, Irpan, Joshi, Julian, Kalashnikov, Kuang, Leal, Lee, Lee, Levine, Lu, Michalewski, Mordatch, Pertsch, Rao, Reymann, Ryoo, Salazar, Sanketi, Sermanet, Singh, Singh, Soricut, Tran, Vanhoucke, Vuong, Wahid, Welker, Wohlhart, Wu, Xia, Xiao, Xu, Xu, Yu, and Zitkovich]{rt22023arxiv}
A.~Brohan, N.~Brown, J.~Carbajal, Y.~Chebotar, X.~Chen, K.~Choromanski, T.~Ding, D.~Driess, A.~Dubey, C.~Finn, P.~Florence, C.~Fu, M.~G. Arenas, K.~Gopalakrishnan, K.~Han, K.~Hausman, A.~Herzog, J.~Hsu, B.~Ichter, A.~Irpan, N.~Joshi, R.~Julian, D.~Kalashnikov, Y.~Kuang, I.~Leal, L.~Lee, T.-W.~E. Lee, S.~Levine, Y.~Lu, H.~Michalewski, I.~Mordatch, K.~Pertsch, K.~Rao, K.~Reymann, M.~Ryoo, G.~Salazar, P.~Sanketi, P.~Sermanet, J.~Singh, A.~Singh, R.~Soricut, H.~Tran, V.~Vanhoucke, Q.~Vuong, A.~Wahid, S.~Welker, P.~Wohlhart, J.~Wu, F.~Xia, T.~Xiao, P.~Xu, S.~Xu, T.~Yu, and B.~Zitkovich.
\newblock Rt-2: Vision-language-action models transfer web knowledge to robotic control.
\newblock In \emph{arXiv preprint arXiv:2307.15818}, 2023.

\bibitem[Cobbe et~al.(2019)Cobbe, Klimov, Hesse, Kim, and Schulman]{cobbe2019quantifyinggeneralizationreinforcementlearning}
K.~Cobbe, O.~Klimov, C.~Hesse, T.~Kim, and J.~Schulman.
\newblock Quantifying generalization in reinforcement learning, 2019.
\newblock URL \url{https://arxiv.org/abs/1812.02341}.

\bibitem[Zhang et~al.(2018)Zhang, Vinyals, Munos, and Bengio]{zhang2018studyoverfittingdeepreinforcement}
C.~Zhang, O.~Vinyals, R.~Munos, and S.~Bengio.
\newblock A study on overfitting in deep reinforcement learning, 2018.
\newblock URL \url{https://arxiv.org/abs/1804.06893}.

\bibitem[Yarats et~al.(2021)Yarats, Fergus, Lazaric, and Pinto]{drqv2}
D.~Yarats, R.~Fergus, A.~Lazaric, and L.~Pinto.
\newblock Mastering visual continuous control: Improved data-augmented reinforcement learning.
\newblock \emph{arXiv preprint arXiv:2107.09645}, 2021.

\bibitem[Lin et~al.(2024)Lin, Hu, Sheng, Wen, You, and Gao]{lin2024data}
F.~Lin, Y.~Hu, P.~Sheng, C.~Wen, J.~You, and Y.~Gao.
\newblock Data scaling laws in imitation learning for robotic manipulation.
\newblock \emph{arXiv preprint arXiv:2410.18647}, 2024.

\bibitem[Lee et~al.(2024)Lee, Wang, Etukuru, Kim, Shafiullah, and Pinto]{lee2024behavior}
S.~Lee, Y.~Wang, H.~Etukuru, H.~J. Kim, N.~M.~M. Shafiullah, and L.~Pinto.
\newblock Behavior generation with latent actions.
\newblock \emph{arXiv preprint arXiv:2403.03181}, 2024.

\bibitem[Zhao et~al.(2023)Zhao, Kumar, Levine, and Finn]{Zhao2023act}
T.~Z. Zhao, V.~Kumar, S.~Levine, and C.~Finn.
\newblock {Learning Fine-Grained Bimanual Manipulation with Low-Cost Hardware}.
\newblock In \emph{Proceedings of Robotics: Science and Systems}, Daegu, Republic of Korea, July 2023.
\newblock \doi{10.15607/RSS.2023.XIX.016}.

\bibitem[Chi et~al.(2023)Chi, Feng, Du, Xu, Cousineau, Burchfiel, and Song]{chi2023diffusionpolicy}
C.~Chi, S.~Feng, Y.~Du, Z.~Xu, E.~Cousineau, B.~Burchfiel, and S.~Song.
\newblock Diffusion policy: Visuomotor policy learning via action diffusion.
\newblock In \emph{Proceedings of Robotics: Science and Systems (RSS)}, 2023.

\bibitem[Chi et~al.(2024)Chi, Xu, Pan, Cousineau, Burchfiel, Feng, Tedrake, and Song]{chi2024universal}
C.~Chi, Z.~Xu, C.~Pan, E.~Cousineau, B.~Burchfiel, S.~Feng, R.~Tedrake, and S.~Song.
\newblock Universal manipulation interface: In-the-wild robot teaching without in-the-wild robots.
\newblock In \emph{Proceedings of Robotics: Science and Systems (RSS)}, 2024.

\bibitem[Li et~al.(2024)Li, Liang, Wang, Luo, Chen, Liao, Wei, Deng, Xu, Zhang, et~al.]{li2024cogact}
Q.~Li, Y.~Liang, Z.~Wang, L.~Luo, X.~Chen, M.~Liao, F.~Wei, Y.~Deng, S.~Xu, Y.~Zhang, et~al.
\newblock Cogact: A foundational vision-language-action model for synergizing cognition and action in robotic manipulation.
\newblock \emph{arXiv preprint arXiv:2411.19650}, 2024.

\bibitem[Wen et~al.(2024)Wen, Zhu, Li, Zhu, Wu, Xu, Cheng, Shen, Peng, Feng, et~al.]{wen2024tinyvla}
J.~Wen, Y.~Zhu, J.~Li, M.~Zhu, K.~Wu, Z.~Xu, R.~Cheng, C.~Shen, Y.~Peng, F.~Feng, et~al.
\newblock Tinyvla: Towards fast, data-efficient vision-language-action models for robotic manipulation.
\newblock \emph{arXiv preprint arXiv:2409.12514}, 2024.

\bibitem[Kroemer et~al.(2021)Kroemer, Niekum, and Konidaris]{kroemer2021review}
O.~Kroemer, S.~Niekum, and G.~Konidaris.
\newblock A review of robot learning for manipulation: Challenges, representations, and algorithms.
\newblock \emph{Journal of machine learning research}, 22\penalty0 (30):\penalty0 1--82, 2021.

\bibitem[Tobin et~al.(2017)Tobin, Fong, Ray, Schneider, Zaremba, and Abbeel]{tobin2017domain}
J.~Tobin, R.~Fong, A.~Ray, J.~Schneider, W.~Zaremba, and P.~Abbeel.
\newblock Domain randomization for transferring deep neural networks from simulation to the real world.
\newblock In \emph{2017 IEEE/RSJ international conference on intelligent robots and systems (IROS)}, pages 23--30. IEEE, 2017.

\bibitem[Andrychowicz et~al.(2020)Andrychowicz, Baker, Chociej, Jozefowicz, McGrew, Pachocki, Petron, Plappert, Powell, Ray, et~al.]{andrychowicz2020learning}
O.~M. Andrychowicz, B.~Baker, M.~Chociej, R.~Jozefowicz, B.~McGrew, J.~Pachocki, A.~Petron, M.~Plappert, G.~Powell, A.~Ray, et~al.
\newblock Learning dexterous in-hand manipulation.
\newblock \emph{The International Journal of Robotics Research}, 39\penalty0 (1):\penalty0 3--20, 2020.

\bibitem[Zhao et~al.(2020)Zhao, Queralta, and Westerlund]{zhao2020sim}
W.~Zhao, J.~P. Queralta, and T.~Westerlund.
\newblock Sim-to-real transfer in deep reinforcement learning for robotics: a survey.
\newblock In \emph{2020 IEEE symposium series on computational intelligence (SSCI)}, pages 737--744. IEEE, 2020.

\bibitem[Haldar et~al.(2023)Haldar, Pari, Rai, and Pinto]{haldar2023fish}
S.~Haldar, J.~Pari, A.~Rai, and L.~Pinto.
\newblock {Teach a Robot to FISH: Versatile Imitation from One Minute of Demonstrations}.
\newblock In \emph{Proceedings of Robotics: Science and Systems}, Daegu, Republic of Korea, July 2023.
\newblock \doi{10.15607/RSS.2023.XIX.009}.

\bibitem[Guzey et~al.(2024)Guzey, Dai, Evans, Chintala, and Pinto]{guzey2024tavi}
I.~Guzey, Y.~Dai, B.~Evans, S.~Chintala, and L.~Pinto.
\newblock See to touch: Learning tactile dexterity through visual incentives.
\newblock In \emph{2024 IEEE International Conference on Robotics and Automation (ICRA)}, pages 13825--13832, 2024.
\newblock \doi{10.1109/ICRA57147.2024.10611407}.

\bibitem[Yuan et~al.(2024)Yuan, Mu, Tao, Fang, Zhang, and Su]{yuan2024policy}
X.~Yuan, T.~Mu, S.~Tao, Y.~Fang, M.~Zhang, and H.~Su.
\newblock Policy decorator: Model-agnostic online refinement for large policy model.
\newblock \emph{arXiv preprint arXiv:2412.13630}, 2024.

\bibitem[Silver et~al.(2018)Silver, Allen, Tenenbaum, and Kaelbling]{silver2018residual}
T.~Silver, K.~Allen, J.~Tenenbaum, and L.~Kaelbling.
\newblock Residual policy learning.
\newblock \emph{arXiv preprint arXiv:1812.06298}, 2018.

\bibitem[Johannink et~al.(2019)Johannink, Bahl, Nair, Luo, Kumar, Loskyll, Ojea, Solowjow, and Levine]{johannink2019residual}
T.~Johannink, S.~Bahl, A.~Nair, J.~Luo, A.~Kumar, M.~Loskyll, J.~A. Ojea, E.~Solowjow, and S.~Levine.
\newblock Residual reinforcement learning for robot control.
\newblock In \emph{2019 International Conference on Robotics and Automation (ICRA)}, pages 6023--6029, 2019.
\newblock \doi{10.1109/ICRA.2019.8794127}.

\bibitem[Zhang et~al.(2019)Zhang, Boehmer, and Whiteson]{zhang2019deep}
S.~Zhang, W.~Boehmer, and S.~Whiteson.
\newblock Deep residual reinforcement learning.
\newblock \emph{arXiv preprint arXiv:1905.01072}, 2019.

\bibitem[Alakuijala et~al.(2021)Alakuijala, Dulac-Arnold, Mairal, Ponce, and Schmid]{alakuijala2021residual}
M.~Alakuijala, G.~Dulac-Arnold, J.~Mairal, J.~Ponce, and C.~Schmid.
\newblock Residual reinforcement learning from demonstrations.
\newblock \emph{arXiv preprint arXiv:2106.08050}, 2021.

\bibitem[Billard and Kragic(2019)]{billard2019trends}
A.~Billard and D.~Kragic.
\newblock Trends and challenges in robot manipulation.
\newblock \emph{Science}, 364\penalty0 (6446):\penalty0 eaat8414, 2019.

\bibitem[Suomalainen et~al.(2022)Suomalainen, Karayiannidis, and Kyrki]{suomalainen2022survey}
M.~Suomalainen, Y.~Karayiannidis, and V.~Kyrki.
\newblock A survey of robot manipulation in contact.
\newblock \emph{Robotics and Autonomous Systems}, 156:\penalty0 104224, 2022.

\bibitem[Hogan et~al.(2020)Hogan, Ballester, Dong, and Rodriguez]{hogan2020tactile}
F.~R. Hogan, J.~Ballester, S.~Dong, and A.~Rodriguez.
\newblock Tactile dexterity: Manipulation primitives with tactile feedback.
\newblock In \emph{2020 IEEE international conference on robotics and automation (ICRA)}, pages 8863--8869. IEEE, 2020.

\bibitem[Whitney and Rourke(1986)]{whitney1986mechanical}
D.~E. Whitney and J.~M. Rourke.
\newblock Mechanical behavior and design equations for elastomer shear pad remote center compliances.
\newblock \emph{Journal of Dynamic Systems, Measurement, and Control}, 108\penalty0 (3):\penalty0 223--232, 09 1986.
\newblock ISSN 0022-0434.
\newblock \doi{10.1115/1.3143771}.
\newblock URL \url{https://doi.org/10.1115/1.3143771}.

\bibitem[Wu et~al.(2024)Wu, Wu, Chen, Chen, Schneider, Johannsmeier, Bing, Abu-Dakka, Knoll, and Haddadin]{wu20241khz}
Y.~Wu, F.~Wu, L.~Chen, K.~Chen, S.~Schneider, L.~Johannsmeier, Z.~Bing, F.~J. Abu-Dakka, A.~Knoll, and S.~Haddadin.
\newblock 1 khz behavior tree for self-adaptable tactile insertion.
\newblock In \emph{2024 IEEE International Conference on Robotics and Automation (ICRA)}, pages 16002--16008. IEEE, 2024.

\bibitem[Pang et~al.(2023)Pang, Suh, Yang, and Tedrake]{pang2023global}
T.~Pang, H.~T. Suh, L.~Yang, and R.~Tedrake.
\newblock Global planning for contact-rich manipulation via local smoothing of quasi-dynamic contact models.
\newblock \emph{IEEE Transactions on robotics}, 39\penalty0 (6):\penalty0 4691--4711, 2023.

\bibitem[Beltran-Hernandez et~al.(2020)Beltran-Hernandez, Petit, Ramirez-Alpizar, and Harada]{beltran2020variable}
C.~C. Beltran-Hernandez, D.~Petit, I.~G. Ramirez-Alpizar, and K.~Harada.
\newblock Variable compliance control for robotic peg-in-hole assembly: A deep-reinforcement-learning approach.
\newblock \emph{Applied Sciences}, 10\penalty0 (19):\penalty0 6923, 2020.

\bibitem[Zhao et~al.(2022)Zhao, Luo, Sushkov, Pevceviciute, Heess, Scholz, Schaal, and Levine]{zhao2022offline}
T.~Z. Zhao, J.~Luo, O.~Sushkov, R.~Pevceviciute, N.~Heess, J.~Scholz, S.~Schaal, and S.~Levine.
\newblock Offline meta-reinforcement learning for industrial insertion.
\newblock In \emph{2022 international conference on robotics and automation (ICRA)}, pages 6386--6393. IEEE, 2022.

\bibitem[Qi et~al.(2023)Qi, Kumar, Calandra, Ma, and Malik]{qi2023hand}
H.~Qi, A.~Kumar, R.~Calandra, Y.~Ma, and J.~Malik.
\newblock In-hand object rotation via rapid motor adaptation.
\newblock In \emph{Conference on Robot Learning}, pages 1722--1732. PMLR, 2023.

\bibitem[Guzey et~al.(2023)Guzey, Evans, Chintala, and Pinto]{guzey2023dexterity}
I.~Guzey, B.~Evans, S.~Chintala, and L.~Pinto.
\newblock Dexterity from touch: Self-supervised pre-training of tactile representations with robotic play.
\newblock \emph{arXiv preprint arXiv:2303.12076}, 2023.

\bibitem[Lin et~al.(2024)Lin, Zhang, Li, Qi, Yi, Levine, and Malik]{lin2024learning}
T.~Lin, Y.~Zhang, Q.~Li, H.~Qi, B.~Yi, S.~Levine, and J.~Malik.
\newblock Learning visuotactile skills with two multifingered hands.
\newblock \emph{arXiv preprint arXiv:2404.16823}, 2024.

\bibitem[Xu et~al.(2023)Xu, Yoon, Fuentes, and Park]{xu2023comprehensive}
M.~Xu, S.~Yoon, A.~Fuentes, and D.~S. Park.
\newblock A comprehensive survey of image augmentation techniques for deep learning.
\newblock \emph{Pattern Recognition}, 137:\penalty0 109347, 2023.
\newblock ISSN 0031-3203.
\newblock \doi{https://doi.org/10.1016/j.patcog.2023.109347}.
\newblock URL \url{https://www.sciencedirect.com/science/article/pii/S0031320323000481}.

\bibitem[Simeonov et~al.(2022)Simeonov, Du, Tagliasacchi, Tenenbaum, Rodriguez, Agrawal, and Sitzmann]{simeonov2022neural}
A.~Simeonov, Y.~Du, A.~Tagliasacchi, J.~B. Tenenbaum, A.~Rodriguez, P.~Agrawal, and V.~Sitzmann.
\newblock Neural descriptor fields: Se (3)-equivariant object representations for manipulation.
\newblock In \emph{2022 International Conference on Robotics and Automation (ICRA)}, pages 6394--6400. IEEE, 2022.

\bibitem[Wang et~al.(2022)Wang, Jia, Zhu, Walters, and Platt]{wang2022robot}
D.~Wang, M.~Jia, X.~Zhu, R.~Walters, and R.~Platt.
\newblock On-robot learning with equivariant models.
\newblock \emph{arXiv preprint arXiv:2203.04923}, 2022.

\bibitem[Florence et~al.(2018)Florence, Manuelli, and Tedrake]{florence2018dense}
P.~R. Florence, L.~Manuelli, and R.~Tedrake.
\newblock Dense object nets: Learning dense visual object descriptors by and for robotic manipulation.
\newblock In \emph{Conference on Robot Learning}, pages 373--385. PMLR, 2018.

\bibitem[Morgan et~al.(2022)Morgan, Hang, Wen, Bekris, and Dollar]{morgan2022complex}
A.~S. Morgan, K.~Hang, B.~Wen, K.~Bekris, and A.~M. Dollar.
\newblock Complex in-hand manipulation via compliance-enabled finger gaiting and multi-modal planning.
\newblock \emph{IEEE Robotics and Automation Letters}, 7\penalty0 (2):\penalty0 4821--4828, 2022.
\newblock \doi{10.1109/LRA.2022.3145961}.

\bibitem[Levy et~al.(2024)Levy, Haldar, Pinto, and Shirivastava]{levy2024p3}
M.~Levy, S.~Haldar, L.~Pinto, and A.~Shirivastava.
\newblock P3-po: Prescriptive point priors for visuo-spatial generalization of robot policies.
\newblock \emph{arXiv preprint arXiv:2412.06784}, 2024.

\bibitem[Haldar and Pinto(2025)]{haldar2025point}
S.~Haldar and L.~Pinto.
\newblock Point policy: Unifying observations and actions with key points for robot manipulation.
\newblock \emph{arXiv preprint arXiv:2502.20391}, 2025.

\bibitem[Mandlekar et~al.(2023)Mandlekar, Nasiriany, Wen, Akinola, Narang, Fan, Zhu, and Fox]{mandlekar2023mimicgendatagenerationscalable}
A.~Mandlekar, S.~Nasiriany, B.~Wen, I.~Akinola, Y.~Narang, L.~Fan, Y.~Zhu, and D.~Fox.
\newblock Mimicgen: A data generation system for scalable robot learning using human demonstrations, 2023.
\newblock URL \url{https://arxiv.org/abs/2310.17596}.

\bibitem[Johns(2021)]{johns2021coarse}
E.~Johns.
\newblock Coarse-to-fine imitation learning: Robot manipulation from a single demonstration.
\newblock In \emph{IEEE International Conference on Robotics and Automation (ICRA)}, 2021.

\bibitem[Iyer et~al.(2024)Iyer, Peng, Dai, Guzey, Haldar, Chintala, and Pinto]{openteach}
A.~Iyer, Z.~Peng, Y.~Dai, I.~Guzey, S.~Haldar, S.~Chintala, and L.~Pinto.
\newblock Open teach: A versatile teleoperation system for robotic manipulation.
\newblock \emph{arXiv preprint arXiv:2403.07870}, 2024.

\bibitem[Teoh et~al.(2024)Teoh, Patidar, Ma, and James]{greenaug}
E.~Teoh, S.~Patidar, X.~Ma, and S.~James.
\newblock Green screen augmentation enables scene generalisation in robotic manipulation.
\newblock \emph{arXiv preprint arXiv:2407.07868}, 2024.

\bibitem[Yuan et~al.(2025)Yuan, Joshi, Zhu, Su, Zhao, and Gao]{roboengine}
C.~Yuan, S.~Joshi, S.~Zhu, H.~Su, H.~Zhao, and Y.~Gao.
\newblock Roboengine: Plug-and-play robot data augmentation with semantic robot segmentation and background generation.
\newblock \emph{arXiv preprint arXiv:2503.18738}, 2025.

\bibitem[Tang et~al.(2023)Tang, Jia, Wang, Phoo, and Hariharan]{dift}
L.~Tang, M.~Jia, Q.~Wang, C.~P. Phoo, and B.~Hariharan.
\newblock Emergent correspondence from image diffusion.
\newblock In \emph{Thirty-seventh Conference on Neural Information Processing Systems}, 2023.

\bibitem[Cheng and Schwing(2022)]{xmem}
H.~K. Cheng and A.~G. Schwing.
\newblock Xmem: Long-term video object segmentation with an atkinson-shiffrin memory model.
\newblock In \emph{European Conference on Computer Vision}, pages 640--658. Springer, 2022.

\bibitem[He et~al.(2016)He, Zhang, Ren, and Sun]{resnet}
K.~He, X.~Zhang, S.~Ren, and J.~Sun.
\newblock Deep residual learning for image recognition.
\newblock In \emph{Proceedings of the IEEE conference on computer vision and pattern recognition}, pages 770--778, 2016.

\bibitem[Bhirangi et~al.(2024)Bhirangi, Pattabiraman, Erciyes, Cao, Hellebrekers, and Pinto]{anyskin}
R.~Bhirangi, V.~Pattabiraman, E.~Erciyes, Y.~Cao, T.~Hellebrekers, and L.~Pinto.
\newblock Anyskin: Plug-and-play skin sensing for robotic touch.
\newblock \emph{arXiv preprint arXiv:2409.08276}, 2024.

\bibitem[Pattabiraman et~al.(2024)Pattabiraman, Cao, Haldar, Pinto, and Bhirangi]{visk}
V.~Pattabiraman, Y.~Cao, S.~Haldar, L.~Pinto, and R.~Bhirangi.
\newblock Learning precise, contact-rich manipulation through uncalibrated tactile skins.
\newblock \emph{arXiv preprint arXiv:2410.17246}, 2024.

\bibitem[Lillicrap et~al.(2015)Lillicrap, Hunt, Pritzel, Heess, Erez, Tassa, Silver, and Wierstra]{ddpg}
T.~P. Lillicrap, J.~J. Hunt, A.~Pritzel, N.~Heess, T.~Erez, Y.~Tassa, D.~Silver, and D.~Wierstra.
\newblock Continuous control with deep reinforcement learning.
\newblock \emph{arXiv preprint arXiv:1509.02971}, 2015.

\bibitem[Smith et~al.(2022)Smith, Kostrikov, and Levine]{smith2022walk}
L.~Smith, I.~Kostrikov, and S.~Levine.
\newblock A walk in the park: Learning to walk in 20 minutes with model-free reinforcement learning.
\newblock \emph{arXiv preprint arXiv:2208.07860}, 2022.

\bibitem[Deitke et~al.(2024)Deitke, Clark, Lee, Tripathi, Yang, Park, Salehi, Muennighoff, Lo, Soldaini, et~al.]{molmo}
M.~Deitke, C.~Clark, S.~Lee, R.~Tripathi, Y.~Yang, J.~S. Park, M.~Salehi, N.~Muennighoff, K.~Lo, L.~Soldaini, et~al.
\newblock Molmo and pixmo: Open weights and open data for state-of-the-art multimodal models.
\newblock \emph{arXiv preprint arXiv:2409.17146}, 2024.

\bibitem[Haldar et~al.(2024)Haldar, Peng, and Pinto]{baku}
S.~Haldar, Z.~Peng, and L.~Pinto.
\newblock Baku: An efficient transformer for multi-task policy learning.
\newblock \emph{arXiv preprint arXiv:2406.07539}, 2024.

\bibitem[Ball et~al.(2023)Ball, Smith, Kostrikov, and Levine]{rlpd}
P.~J. Ball, L.~Smith, I.~Kostrikov, and S.~Levine.
\newblock Efficient online reinforcement learning with offline data.
\newblock In \emph{International Conference on Machine Learning}, pages 1577--1594. PMLR, 2023.

\bibitem[Karpathy(2021)]{minGPT}
A.~Karpathy.
\newblock mingpt: A minimal pytorch re-implementation of the openai gpt.
\newblock \url{https://github.com/karpathy/minGPT}, 2021.

\end{thebibliography}

\newpage
\appendix
\section*{Appendix}
\renewcommand{\thesubsection}{A\arabic{subsection}}
\subsection{Hyperparameters and Network Architecture}
\label{app:hyperparams}

    \subsubsection{Hyperparameters} The complete list of hyperparameters is provided in Table~\ref{table:hyperparams}. We collect 32 demonstrations for each task using a VR-based teleoperation framework~\cite{openteach} operating at 30Hz. The collected data is subsampled to 6Hz for policy training, and the learned policies are deployed at 6Hz during real-world execution. All training is done on a local desktop with an NVIDIA RTX 3080 GPU with 8GB VRAM. For BAKU~\cite{baku}, ViSk~\cite{visk}, and \methodbc{}, training for 20k iterations with a training time of around 30 minutes provided the best results. For online RL (RLPD~\cite{rlpd}, \method{}), each method is trained online for 45 minutes at a 6Hz frequency(around 16k environment steps). For both offline and online training, the image observations are augmented with random cropping and color jitter. For tactile observations from the AnySkin~\cite{anyskin} sensor, we subtract a baseline measurement from each tactile reading to account for sensor drift~\cite{visk}.

\subsubsection{Network Architecture} We use a randomly initialized ResNet-18~\cite{resnet} as our image encoder and a 2-layer MLP for encoding the 15-dimensional tactile reading from AnySkin. For offline training, we use a transformer-based policy~\cite{baku,visk}, using the minGPT~\cite{minGPT} architecture for the transformer. For offline training, all baselines and \method{} use an MLP action head, predicting a chunk of 10 future actions. Instead of using only the current action prediction, we use a temporal ensemble to combine all the past chunked action predictions. This temporal ensemble performs a weighted average over these predictions with an exponential weighing scheme $w_{i} = exp(-m * i)$, where $w_0$ is the weight for the oldest action. The speed for incorporating a new observation is governed by $m$, where a smaller $m$ means faster incorporation. It must be noted that this ensembling incurs no additional training cost, only extra inference-time computation. In our experiments, similar to prior work~\cite{baku,visk}, we find both action chunking and temporal ensembling to be important for producing precise and smooth motion. During online residual RL, an offset is learned on top of the temporally smoothed offline action. We set $m$ to 0.01 for all our experiments. 

During online residual RL, the offset scale is chosen to balance exploration capacity about the base action and the convergence speed of online training. Since we focus on precise tasks demanding sub-millimeter level accuracy, we set to maximum offset magnitude to be 20\% of the maximum action observed in the training data. Further, we observed that exploration plays an important role in preventing early collapse during online training. Thus, we employ a linearly decaying standard-deviation schedule -- starting with high noise during the initial RL phase to ensure flexibility, then gradually reducing it to guarantee stable convergence. For residual RL, the actor is a 1-layer MLP while the critic comprises 2-MLP layers. For RLPD, which trains the RL policy from scratch, we use a 4-layer MLP for the actor network.

\methodbc{} has a total of 7.6M parameters, while \method{} has an additional 2.46M from the residual RL phase, resulting in a total of 9.06M parameters.


\begin{table}[h]
    \caption{List of hyperparameters.}
    \label{table:hyperparams}
    \begin{center}
    \setlength{\tabcolsep}{18pt}
    \renewcommand{\arraystretch}{1.5}
    \begin{tabular}{ l l } 
        \toprule
        Parameter                  & Value \\
        \midrule
        Learning rate              & $1e^{-4}$\\
        Image size                 & $128\times 128$ \\
        Batch size                 & 256 \\
        Optimizer                  & Adam\\
        Hidden dim                 & 256\\
        Observation history length & 1 \\
        Action head                & MLP \\
        Action chunk length        & 10\\ 
        Residual RL Offset scale   & 20\% of max absolute action \\
        Exploration schedule for RL  & linear(0.25,0.1,5000)\\
        Update-to-data ratio (UTD) & 16
\\        \bottomrule
        
    \end{tabular}
    \end{center}
\end{table}

\subsection{Task Descriptions}
\label{app:task-descriptions}

\noindent\textbf{Plug in Socket} The robot arm holds a plug within the gripper and is tasked with inserting the plug into a socket. The robot's initial position is randomly sampled in a 6cm$\times$6cm area around the socket, 10cm above the socket.

\noindent\textbf{USB Insertion} The robot arm holds a USB stick within the gripper and is tasked with going down and inserting the USB stick into a USB socket. The robot's initial position is randomly sampled in a 6cm$\times$6cm area around the socket, 10cm above the socket.

\noindent\textbf{Card Swiping} The robot arm holds a credit card within the gripper and is tasked with swiping the card through a card machine. The robot's initial position is sampled in a 4cm$\times$4cm$\times$2cm area in front of the card machine.

\noindent\textbf{Key in Lock} The robot arm holds a key within the gripper and is tasked with inserting the key into a lock. The robot's initial position is sampled in a 6cm$\times$6cm area around the key hole, 10 cm above the socket.

\noindent\textbf{Pick block} The robot arm is tasked with picking up a block placed at a fixed position on the table. The robot's initial position is sampled in a 6cm$\times$6cm area around the block, 10 cm above the block.

\subsection{Baseline Implementations}
\label{app:baseline}

\paragraph{BAKU~\cite{baku}} This is a visual behavior cloning baseline using a transformer architecture and a deterministic MLP action head. We follow the hyperparameters and network architectures described in Appendix~\ref{app:hyperparams} for BAKU.

\paragraph{ViSk~\cite{visk}} This is a visuo-tactile behavior cloning baseline using a transformer architecture and a deterministic MLP action head. We follow the hyperparameters and network architectures described in Appendix~\ref{app:hyperparams} for ViSk.

\paragraph{RLPD~\cite{rlpd}} This involves collecting a few expert demonstrations and training an RL policy from scratch, where the data during RL training is sampled 1:1 between the expert and RL replay buffer. RLPD employs a high update-to-data ratio (UTD) and layer normalization in the critic to enable sample-efficient online learning. We observe that during the initial phase of training, since the actor is randomly initialized, the RLPD policy outputs unsafe actions, making the rollout jerky. This highlights the importance of pretraining for stable online learning.


\begin{figure}[t]
    \centering
    \includegraphics[width=\linewidth]{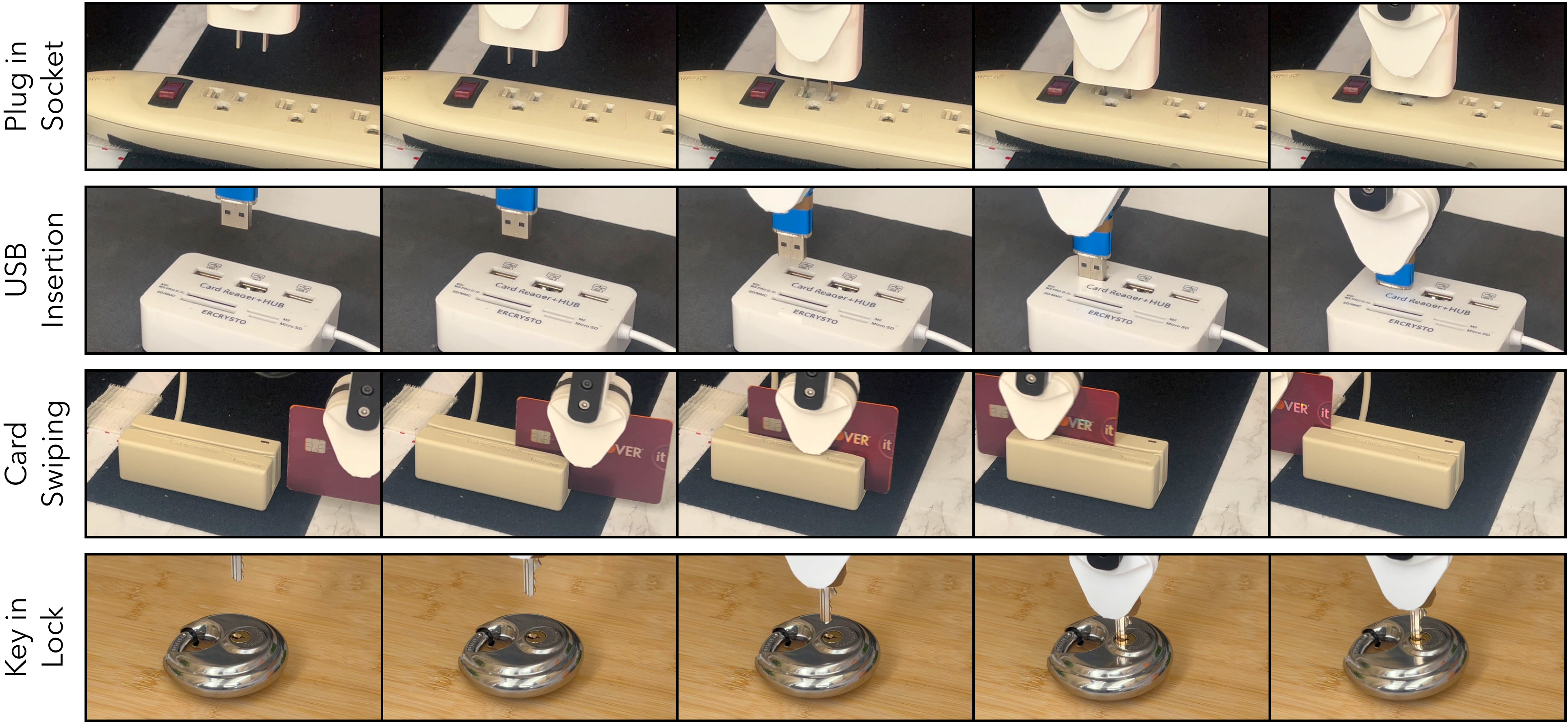}
    \caption{Real-world rollouts showing \method{}'s ability on four precise manipulation tasks.}
    \label{fig:rollouts}
\end{figure}

\begin{figure}[t]
    \centering
    \includegraphics[width=\linewidth]{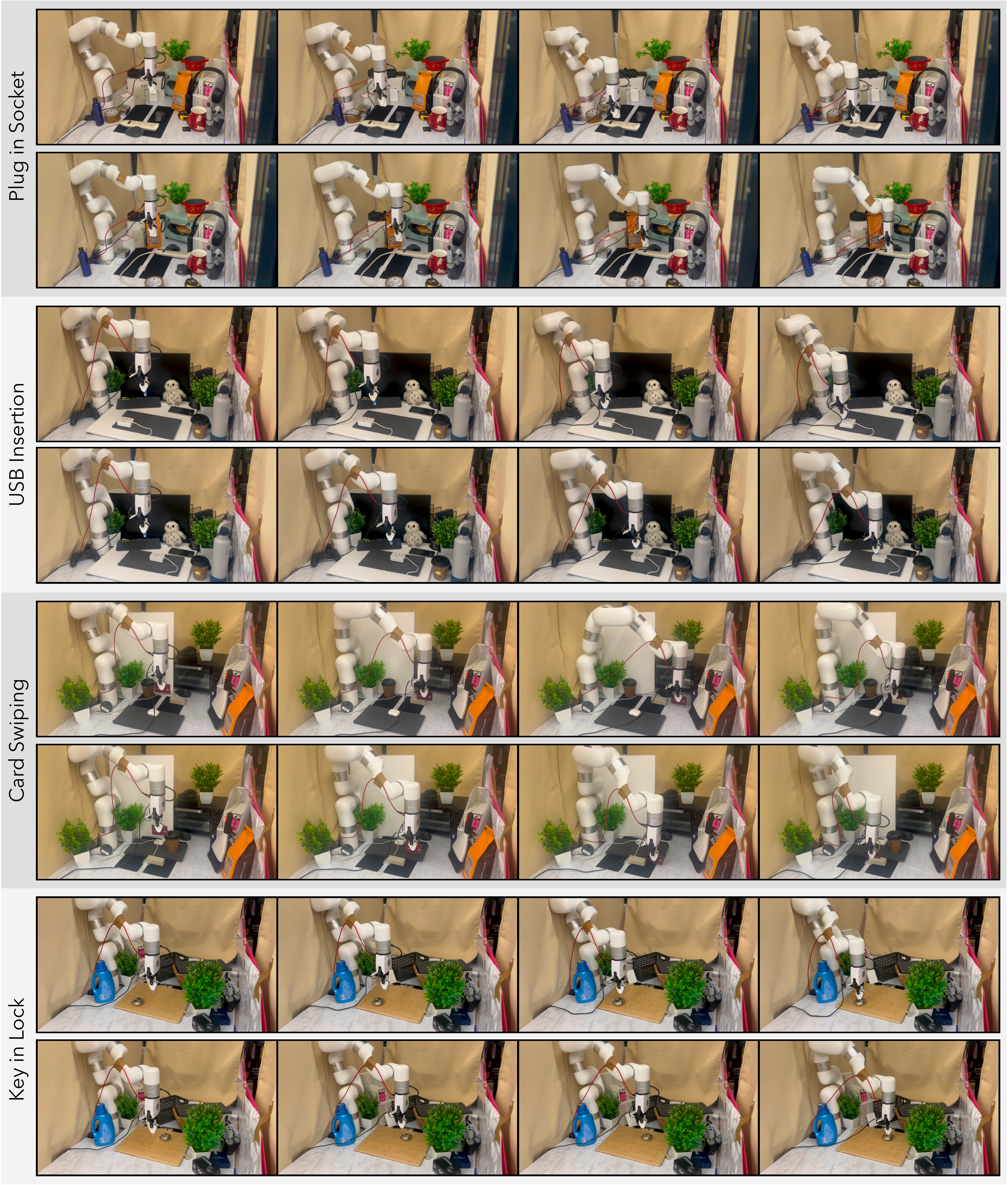}
    \caption{Real-world rollouts showing that \method{} generalizes to spatial variations and background distractor objects.}
    \label{fig:generalization}
\end{figure}

\subsection{Spatial and Scene Generalization}
\label{app:scenegen}

Figure~\ref{fig:rollouts} demonstrates \method{} performing four precise manipulation tasks in the same position as during training, but with a novel background. Figure~\ref{fig:generalization} further highlights \method{}'s spatial and scene generalization capabilities: spatial generalization is achieved through VLM-guided reaching in conjunction with a localized observation and action space, while semantic augmentations support scene generalization during training, which promote invariance to changes in background, textures, lighting, and clutter. Together, this allows \method{} to decouple spatial reasoning from scene understanding, enabling reliable, precise manipulation across a wide range of previously unseen environments without retraining.

\end{document}